%% file: bare_jrnl.tex
\documentclass[journal]{IEEEtran}%

\usepackage{amsmath,amssymb} 
\usepackage{cite}
\usepackage{url}
\usepackage{graphicx}
\usepackage[usenames,dvipsnames]{color}
\usepackage{array}
\usepackage{algorithm,algorithmic}
\usepackage{setspace}
\usepackage{multirow}
\usepackage{multicol}
\usepackage{mathtools}
\usepackage{amsthm}
\usepackage{lipsum}
\usepackage{dblfloatfix}
\usepackage{tabulary}

\newcounter{tempEquationCounter} 
\newcounter{thisEquationNumber}

\newcommand{\norm}[1]{\left\lVert#1\right\rVert}

\newtheorem{theorem}{Special Theorem}
\newtheorem{lemma}[theorem]{Lemma}
\DeclareMathOperator*{\argmin}{\arg\!\min}

\newcommand\scalemath[2]{\scalebox{#1}{\mbox{\ensuremath{\displaystyle #2}}}}

\definecolor{light-gray}{gray}{0.4}

\def\ontop#1#2{\setbox0\hbox{#2}\copy0\llap{\raise\ht0\hbox{#1}}}

\begin{document}
%
\title{Motion Estimation via Robust Decomposition with Constrained Rank}
%
%
%

\author{German~Ros*, Jose~M.~\'Alvarez,
        and~Julio~Guerrero
\thanks{G. Ros is with the Computer Vision Center at Universitat Aut\`onoma de Barcelona, Spain, e-mail: gros@cvc.uab.es .}
\IEEEcompsocitemizethanks{\IEEEcompsocthanksitem Jose~M.~\'Alvarez is a researcher at NICTA, 2601 Canberra, Australia, e-mail: Jose.Alvarez@nicta.com.au.}%
\thanks{J. Guerrero is with the Department of Applied Mathematics at Universidad de Murcia, Spain, e-mail: juguerre@um.es .}%
}%



%
%

\markboth{\LaTeXe Journal}%
{Ros et.al \MakeLowercase{\textit{et al.}}: Motion Estimation via Robust Decomposition with Constrained Rank}
%



\maketitle

%
\IEEEpeerreviewmaketitle

\input{sections/Abstract}

\input{sections/Introduction}
\input{sections/RelatedWork}
\input{sections/Problem}
\input{sections/Approach}
\input{sections/Results}
\input{sections/Conclusions}
\input{sections/Ack}

\bibliographystyle{IEEEtran}
\bibliography{sections/paperbib}

%


\vspace{-14mm}
\begin{IEEEbiography}[{\includegraphics[width=1in,height=1.25in,clip,keepaspectratio]{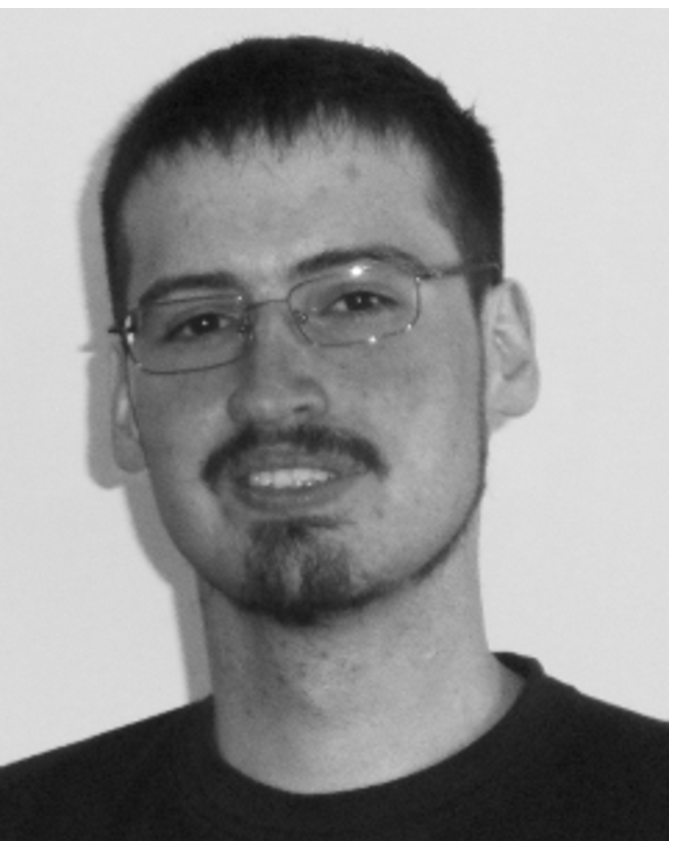}}]{German Ros}
received his B.Sc (Hons.) in computer science from Universidad de Murcia (Spain) in 2010 and his M.Sc degree from Kingston University of London (UK) in 2011. In 2012 he obtained a second M.Sc degree from Universitat Aut\`onoma de Barcelona, where he is currently pursuing a Ph.D. His research interests span computer vision, robotics and visual perception, with special interest in manifold optimization, robust estimation and visual geometry. He is a student member of the IEEE.
\end{IEEEbiography}

\vspace{-16mm}

\begin{IEEEbiography}[{\includegraphics[width=1in,height=1.25in,clip,keepaspectratio]{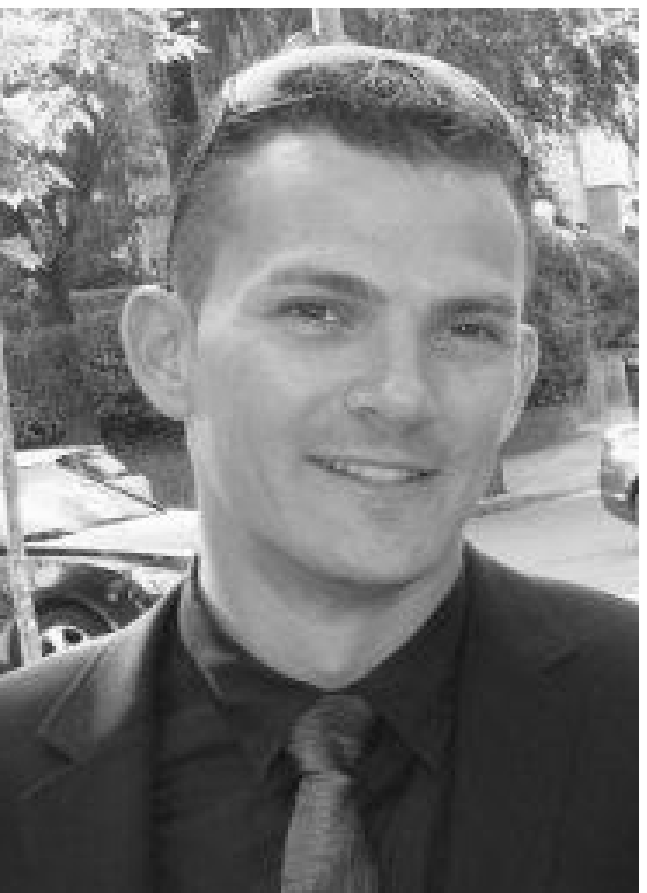}}]{Jose~M.~\'Alvarez}
is currently a senior researcher at NICTA and a research fellow at the Australian National University.
Previously, he was a postdoctoral researcher at the Computational and Biological Learning
Group at New York University. During his Ph.D. he was a visiting researcher at the University of Amsterdam and VolksWagen research. His main research interests include road detection, color, photometric invariance, machine learning, and fusion of classifiers. He is associate editor for the IEEE Trans. on Intelligent Transportation Systems and member of the IEEE.\vspace{-1.0cm}
\end{IEEEbiography}

\vspace{-6mm}

\begin{IEEEbiography}[{\includegraphics[width=1in,height=1.25in,clip,keepaspectratio]{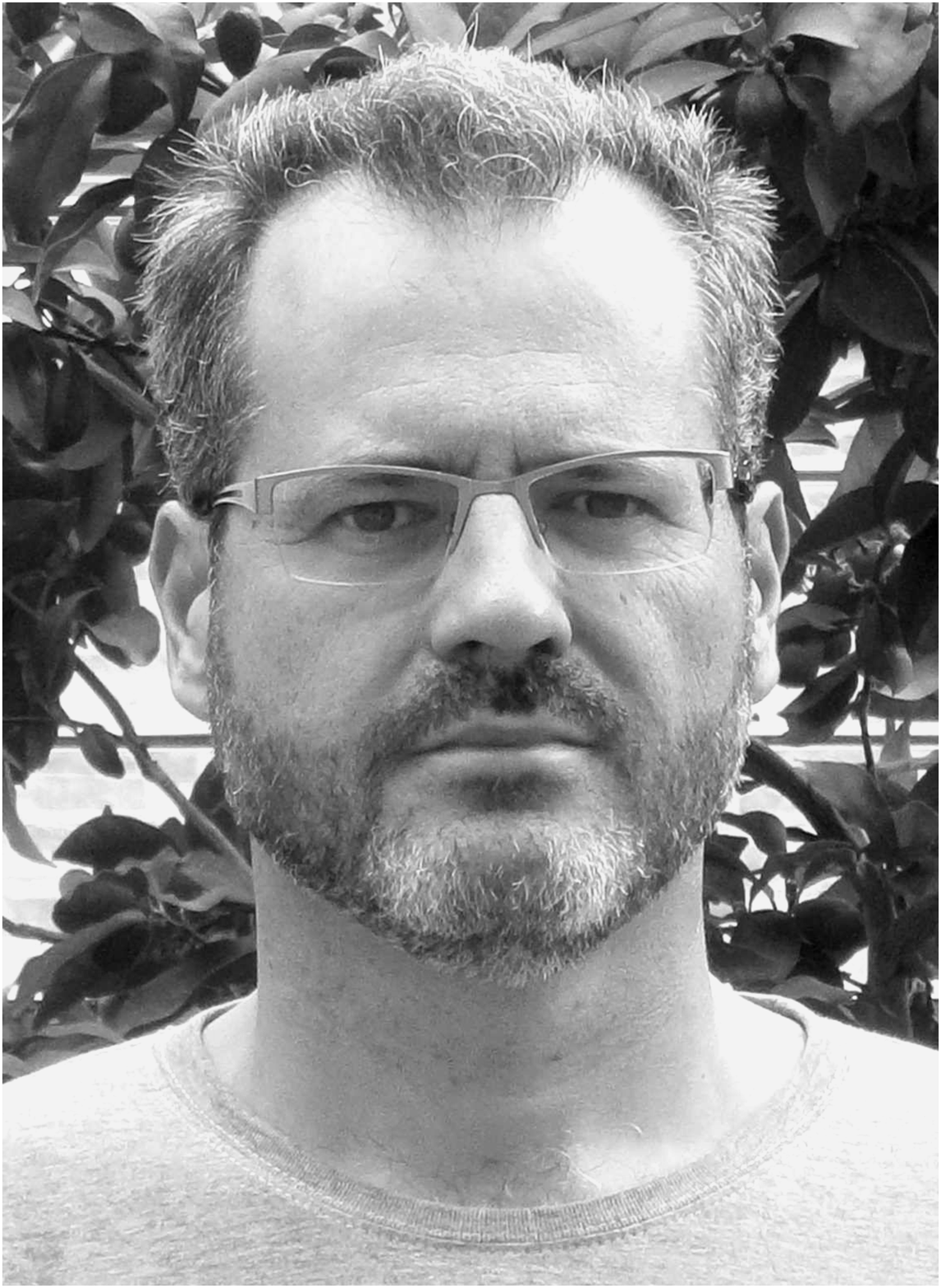}}]{Julio Guerrero}
 is an Associate Professor in the Department of Applied Maths  at Universidad de Murcia (Spain) since 2000. Previously he was a postdoctoral researcher at Naples University, and visiting researcher at Durhan University (UK) and Syracure Univerity (USA). His research interests include, among others, applied maths and computer vision, with special interest in manifold optimization and robust estimation.
\end{IEEEbiography}





\end{document}

%% file: sections/Abstract.tex
\begin{abstract}

In this work, we address the problem of outlier detection for robust motion estimation by using modern sparse-low-rank decompositions, i.e., Robust PCA-like methods, to impose global rank constraints. Robust decompositions have shown to be good at splitting a corrupted matrix into an uncorrupted low-rank matrix and a sparse matrix, containing outliers. However, this process only works when  matrices have relatively low rank with respect to their ambient space, a property not met in motion estimation problems. As a solution, we propose to exploit the partial information present in the decomposition to decide which matches are outliers. We provide evidences showing that even when it is not possible to recover an uncorrupted low-rank matrix, the resulting information can be exploited for outlier detection. To this end we propose the Robust Decomposition with Constrained Rank (RD-CR), a proximal gradient based method that enforces the rank constraints inherent to motion estimation. We also present a general framework to perform robust estimation for stereo Visual Odometry, based on our RD-CR and a simple but effective compressed optimization method that  achieves high performance. Our evaluation on synthetic data and on the KITTI dataset demonstrates the applicability of our approach in complex scenarios and it yields state-of-the-art performance.
\end{abstract}


%% file: sections/Introduction.tex
\section{Introduction}

\IEEEPARstart{T}{his} paper addresses the application of robust decomposition techniques, like the successful Robust PCA (RPCA)~\cite{Wright09}\cite{Candes11}, to the problem of outlier (i.e., gross noise) detection in the special case of camera-pose recovery and Visual Odometry, referred in  this paper to as motion estimation. 

\begin{figure*}[!t]
	\centering
	\includegraphics[scale=0.80]{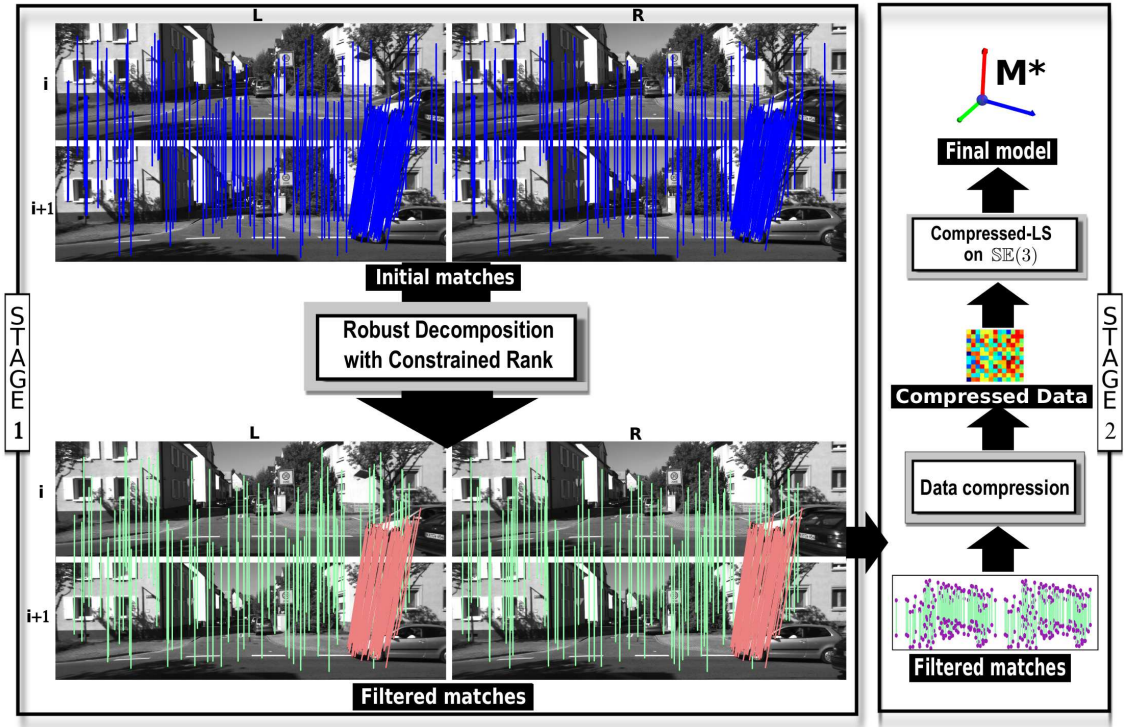}
	\vspace{-2mm}
	\caption{Proposed approach as a two stages pipeline, i.e.: RD-CR for outlier detection followed by a stage of data reduction and optimization. Notice how the method can correctly identify the matches from the moving car as outliers during stage one.}
	\label{fig:pipeline}
	\vspace{-4mm}
\end{figure*}

The emergence of outliers in real situations is unavoidable and estimation techniques need to achieve a certain level of robustness to prevent producing wrong estimations. In the context of motion estimation ---the practical case studied in this paper--- outliers arise by two main reasons. The first reason is the violation of model assumptions, for instance, when a scene is considered static but foreground objects start moving (see the car in Fig.~\ref{fig:pipeline}). The second is the failure of matching methods to retrieve correct point matches at different viewpoints.
 
Recently, it has been shown that RPCA can be used to recover an uncorrupted version of a matrix affected by outliers~\cite{Wright09}\cite{Candes11}\cite{RPCAOP}, but such a recovery is just achievable when matrices fulfil several well-understood constraints~\cite{Universality} (see Section~\ref{sec:problem}). However, this is not the case when dealing with the data matrices arising in motion estimation problems, and therefore RPCA can not be used to exactly recover an uncorrupted version of the data, i.e., such matrices are outside the recovery boundaries of RPCA~\cite{Wright09}\cite{Candes11}.

In this paper we propose to exploit the information resultant from the  sparse and low-rank decomposition to make a binary decision on each tuple of point matches (column) about its pertinence to the outlier set. As we will show along this paper, despite the impossibility of performing an exact recovery of every element of the observation matrix, the resultant information is enough to make this set of binary decisions. Unlike works like~\cite{RPCAOP}, our outlier detection technique has the advantage that is not based on the correct recovery of the low-rank matrix and the sparse matrix (or their column spaces). Instead, we exploit the rank constraints present in motion estimation problems, leading to a novel Robust Decomposition with Constrained Rank (RD-CR). Our technique is based on proximal gradients, inspired by the Accelerated Proximal Gradient algorithm (APG)~\cite{APG}, but extended to deal with fixed-rank constraints efficiently. We show that our approach presents extended boundaries of application, making robust decomposition techniques useful for motion estimation applications. 

We also present a new framework for robust motion estimation that applies RD-CR as a preprocessing stage, prior to the application of a very fast compressed least-squares method. Compressed least-squares makes use of an efficient data compression scheme known as the Reduced Measurement Matrix (RMM)~\cite{AlgError}\cite{Ros2013b} to condense thousands of constraints (point matches) into a compact matrix surrogate that allows for very fast optimization. RD-CR mitigates the robustness problem of compressed least-squares methods by filtering out the outliers. This finally leads to an efficient and robust regression framework. We demonstrate this framework for the stereo Visual Odometry problem with evaluations in both synthetic cases and the realistic and challenging sequences of the KITTI dataset~\cite{KITTI}. This evaluation also serves to test the applicability of RD-CR technique in complex scenarios, showing its good virtues when compared against state-of-the-art techniques. To the best of our knowledge this is the first time robust decompositions are tested in such challenging scenarios.

%% file: sections/RelatedWork.tex
\vspace{-1mm}
\section{Related Work}
\label{sec:related}

\subsection{Robust Estimation}

Robust estimation aims at providing algorithms with certain level of resistance to noise and outliers~\cite{ComparativeAnalysisRANSAC}\cite{RobustStatistics}. It is a well-studied problem within the field of statistics and plays a crucial role in computer vision and robotics. Within computer vision and robotics, these techniques are widely applied to motion estimation problems, such as camera pose estimation and Visual Odometry (VO). In these domains, state-of-the-art approaches are usually based on consensus strategies, like the RANdom SAmple Consensus (RANSAC)\cite{Ransac} or any of its variants~\cite{ComparativeAnalysisRANSAC}. The main problem of RANSAC is its computational complexity. RANSAC estimates robust models by identifying inliers via the generation of several hypotheses (models). This process grows exponentially w.r.t the percentage of outliers, leading to undesired performance for practical applications. A recent variation of consensus methods, called robust averaging~\cite{LieAlgebraicAvg}\cite{L1rotation}, avoids this drawback by merging several ``weak'' models in a robust fashion to produce a robust outcome. In this case less models are needed, leading to a more efficient approach. However these methods have a breakdown point of $\beta=0.5$.

In many occasions, simpler alternatives, such as robust cost functions, are required to perform a quick and simple job. An instance of this is the Huber M-estimator~\cite{Huber}\cite{RobustStatistics}, a variation of least-squares that uses two different modes to weight data points close to zero (inliers) and those that are far away (outliers) . Unfortunately, although efficient, its theoretical breakdown point is $\beta = 0$, meaning that in the worst case a single gross error could produce an arbitrarily large bias in the estimation. Another example is the Iteratively Reweighted Least Squares (IRLS) technique~\cite{Holland77}\cite{Leary90}. In this case robustness is achieved by dynamically modifying a set of weights describing the influence of each observation. When this update is based on a robust cost like the $\ell_1$-norm (IRLS-L1) the breakdown point becomes $\beta=0.5$, therefore, being a good alternative to consensus approaches~\cite{Holland77}. 

Achieving a breakdown point of $\beta = 0.5$ is the best one can expect without any extra information of the data distribution. In previous cases, all the methods try to detect outliers by analysing how data samples agree with a set of putative models. This exhaustive generation leads to an expensive process. A promising alternative we use in this paper consists in analysing the quality of the data with respect to an abstract set of constraints, such as the rank constraints arising in $\mathbb{SO}(N)$, $\mathbb{SE}(N)$ groups. These rank constraints can be exploited very efficiently without the need of creating a set of putative models. Instead, observations are checked as a ``whole'' against the rank constraint to decide which elements are more likely to violate the condition. Moreover, these properties still hold beyond the breakdown point $\beta=0.5$, at least theoretically. 

\subsection{RPCA for Outlier Detection}

Outlier detection has been previously addressed by RPCA-like techniques in different domains, such as identifying outliers in digits images~\cite{RPCAOP}. However, in these cases the ambient dimension is high and the rank of the data matrix is low (i.e., the recovery warranties are met). As a consequence, outlier detection is posed as the problem of recovering a clean column space of the input data by mean of a robust decomposition based on a $\ell_{1,2}$ penalizer. However, the problem of motion estimation from stereo devices involves dealing with a low ambient dimension problem, which is very challenging due to theoretical limitations. 

An initial attempt to deal with this problem can be found in~\cite{SOLO12}. The proposed algorithm performs outlier detection for 3D registration by using a RPCA-like approach. However, they always stay within the recovery boundaries by gathering as many frames as possible, creating a high ambient dimension. This avoids the main problem of the low ambient dimension but at the cost of limiting the applicability of the technique, which can only work in small scenarios where conditions remain similar through time.

In this paper we propose to overcome these limitations with a more general alternative that is suitable for large sequences in outdoor environments, where only pairs of frames are required. The recovery boundaries are not fulfilled, but even in these extreme situations, our approach produces useful information that helps detecting outliers. Moreover, we assume that our models are based on multiple 2D observations~\cite{VastScale}\cite{VisualOdometryStereo} instead of 3D. This has proven to be more convenient for Visual Odometry, where, least-squares based methods are frequently used to model the problem and the Maximum Likelihood Estimator (MLE) of these functions is achieved by using 2D observations.

%% file: sections/Problem.tex
\section{Problem formulation}
\label{sec:problem}

In this work, we consider the estimation of motion models $\{M_i\}_{i=1}^{N-1}$, between pair of stereo frames $F_i \rightarrow F_{i+1}$ along a given sequence of $N$ frames $\{F_i\}_{i=1}^{N}$. Each frame $F_i = (V_i^l, V_i^r)$ consists of two images taken from the left and right cameras at the same time instant $t_i$. This formulation is suitable for the stereo Visual Odometry problem addressed in this work. To this end, we gather observations from two cameras in two different time instants. Camera models are represented by $C_i^l$, $C_i^r, C_{i+1}^l, C_{i+1}^r$, where superindices stand for (l)eft and (r)ight cameras and subindices denote time instants. Without lack of generality, we assume that the stereo rig is calibrated and that the image planes of both cameras, $C^l_*$ and $C^r_*$, are aligned and have a known baseline $B$. The motion of the stereo rig from time $t_i$ to $t_{i+1}$ is given by a rigid 3D transformation $M_i$ as shown in (\ref{eq:mo}),

\begin{equation}
\label{eq:mo}
 (C_i^l, C_i^r) \overset{M_i \in \mathbb{SE}(3)}{\longrightarrow} (C_{i+1}^l, C_{i+1}^r).
\end{equation}

This motion is represented as an element of the Special Euclidean group in 3D as in 

\begin{equation}
\label{eq:se3}
M_i \in \mathbb{SE}(3) = {\left[ \begin{smallmatrix}R_{3 \times 3} & T_{3 \times 1}\\0_{1 \times 3} & 1\end{smallmatrix} \right]},
\end{equation}

\noindent where $R$ is a rotation matrix and $T$ a translation vector. Each of the cameras can be algebraically described as shown in (\ref{eq:cameras2})--(\ref{eq:cameras3}),

\begin{align}
\label{eq:cameras2}
C^l_i &= \Pi K \left[\begin{smallmatrix} I_{3 \times 3} & 0_{3 \times 1}\\ 0_{1 \times 3} & 1\end{smallmatrix}\right]\\
C^l_{i+1} &= \Pi K \left[\begin{smallmatrix} I_{3 \times 3} & 0_{3 \times 1}\\ 0_{1 \times 3} & 1\end{smallmatrix}\right] M_i\\
C^r_i &= \Pi K \left[\begin{smallmatrix} I_{3 \times 3} & -B_{3 \times 1}\\ 0_{1 \times 3} & 1\end{smallmatrix}\right]\\
C^r_{i+1} &= \Pi K  \left[\begin{smallmatrix} I_{3 \times 3} & -B_{3 \times 1}\\ 0_{1 \times 3} & 1\end{smallmatrix}\right] M_i,
\label{eq:cameras3}
\end{align}

\noindent given that $\Pi(\cdot) : \mathbb{R}^4\rightarrow\mathbb{R}^3$, represents a projection matrix, and 

\begin{equation}
K = \scalemath{1}{\left[ \begin{smallmatrix}f & 0 & c_u & 0\\ 0 & f & c_v & 0\\ 0 & 0 &1 & 0 \\ 0 & 0 & 0 & 1\end{smallmatrix} \right]},
\end{equation}

\noindent is a calibration matrix with focal length $f$ and principal point at $(c_u, c_v)$. Given this configuration, the transformation $M_i$ of the stereo rig can be estimated from a set of $N_c$ 2D correspondences $\{(x_i^l \leftrightarrow x_i^r \leftrightarrow x_{i+1}^l \leftrightarrow x_{i+1}^r)^{(j)}\}_{j=1}^{N_c}$ coming from the views $V_i^l, V_i^r, V_{i+1}^l, V_{i+1}^r$, respectively~\cite{VastScale}\cite{VONyster}\cite{DWO}. When estimating the transformation $M_i$, one should account for the presence of noise and outliers in the observations in order to avoid a biased solution. 

Our proposal exploits the rank constraints present in rigid 3D motions to identify outliers. To achieve this, it is important to notice that some arrangements of the input data as a matrix lead to such rank constraints~\cite{SOLO12}\cite{Tomasi92}. As an example of this concept, let us consider two sets of homogeneous 3D points at two time instants $\mathcal{X}_i, \mathcal{X}_{i+1}$, stacked as described in (\ref{eq:rank})-left. When the right side expresses a matrix in terms of an initial set of points and two motion transformations acting on it, as

\begin{equation}
\underbrace{{\left[\begin{smallmatrix}X_i^1 &  \cdots & X_i^{N_c}\\ X_{i+1}^1 &  \cdots & X_{i+1}^{N_c}\end{smallmatrix}\right]}\llap{\lower +3ex\hbox{$\scriptscriptstyle 8 \times N_c$}\hskip+5ex}}_{\text{rank} \leq 4} = \underbrace{{\left[\begin{smallmatrix}M_{i_{3 \times 4}} \\ M_{j_{3 \times 4}}\end{smallmatrix}\right]}\llap{\lower +3ex\hbox{$\scriptscriptstyle 8 \times 4$}\hskip+2ex} }_{\text{rank} \leq 4} \underbrace{{\left[\begin{smallmatrix} X_i^1 & \dots & X_i^{N_c} \end{smallmatrix}\right]}\llap{\lower +3ex\hbox{$\scriptscriptstyle 4 \times N_c$}\hskip+4ex}}_{\text{rank} \leq 4}.
\label{eq:rank}
\end{equation}

\begin{figure}[!t]
	\centering
	\includegraphics[scale=0.65]{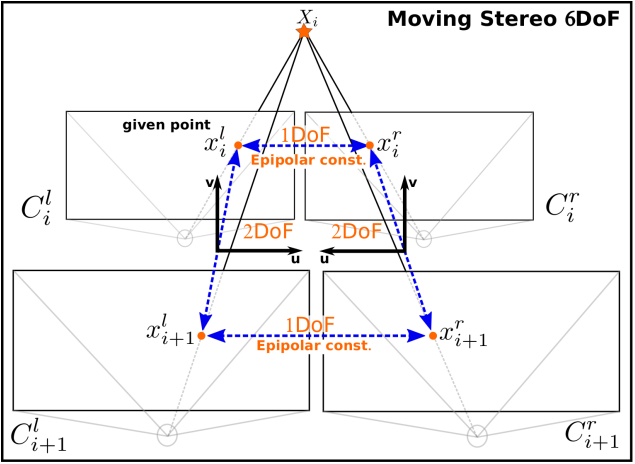}
	\vspace{-3mm}
	\caption{Diagram representing the geometrical configuration of a moving stereo camera. We show a 3D point $X_i$ projected onto the four views and the arising six degrees of freedom.}
	\label{fig:proof}
	\vspace{-4mm}
\end{figure}

From this expression is easy to see that the rank of the matrix is $r \leq 4$, since $\text{rank}(A B) \leq min(\text{rank}(A), \text{rank}(B))$. This example shows the rank constraints exploited by our approach. However, in our case we use multi-view 2D observations from a calibrated stereo rig, instead of 3D points. The reasoning for this case, is given in Lemma~\ref{prop1} and Fig.~\ref{fig:proof}. 

\vspace{1mm}\begin{lemma}\label{prop1} Let us assume a measurement matrix $W$ made of uncorrupted 2D observations that have been normalized to remove their mean or transformed according to $K^{-1}$ (known). $W$ consists of observations from four views of a moving stereo rig. Let also consider without lack of generality that the planes of the left and right cameras are aligned and that the separation $B$ between the cameras forming the stereo rig defined in one direction (x-axis).  In these conditions, the 3D motion $M_i \in \mathbb{SE}(3)$ of the stereo rig will lead to $\text{rank}(W) \leq 6$ as expressed in (\ref{eq:rank2}).  

\vspace{-5mm}
\begin{equation}
\scalemath{0.95}{ W \hskip-.4ex = \hskip-.6ex \underbrace{\underset{
 8 \times N_c}{\left[\begin{smallmatrix}
x^{l, 1}_i  & \dots & x^{l, N_c}_i\\
x^{r, 1}_i  & \dots & x^{r, N_c}_i\\
x^{l, 1}_{i+1}  & \dots & x^{l, N_c}_{i+1}\\
x^{r, 1}_{i+1}  & \dots & x^{r, N_c}_{i+1}
\end{smallmatrix}\right]}}_{\text{rank} \leq 6} \hskip-.5ex = \hskip-.7ex h^{-1} \hskip-.7ex
 \left( \hskip-1ex \underset{12 \times 4}{\vphantom{{\left[\begin{smallmatrix}
x^{l, 1}_i  & \dots & x^{l, N_c}_i\\
x^{r, 1}_i  & \dots & x^{r, N_c}_i\\
x^{l, 1}_{i+1}  & \dots & x^{l, N_c}_{i+1}\\
x^{r, 1}_{i+1}  & \dots & x^{r, N_c}_{i+1}
\end{smallmatrix}\right]}}  \left[\begin{smallmatrix} C^l_i\\C^r_i\\C^l_{i+1}\\C^r_{i+1}\end{smallmatrix}\right]} \underset{4 \times N_c}{ \vphantom{{\left[\begin{smallmatrix}
x^{l, 1}_i  & \dots & x^{l, N_c}_i\\
x^{r, 1}_i  & \dots & x^{r, N_c}_i\\
x^{l, 1}_{i+1}  & \dots & x^{l, N_c}_{i+1}\\
x^{r, 1}_{i+1}  & \dots & x^{r, N_c}_{i+1}
\end{smallmatrix}\right]}} \hskip-.7ex \left[\begin{smallmatrix} X_i^{l,1} & \cdots & X_i^{l,N_c} \end{smallmatrix}\right]} \hskip-.9ex \right).
}
\label{eq:rank2}
\end{equation}
\vspace{-4mm}
\end{lemma}

Here, $h^{-1}(\cdot)$ is a non-linear function that converts each point from homogeneous to non-homogeneous coordinates.  This operation rises the maximum rank from 4, the case explained before for 3D data, to 6. In the following section we explain how to perform outlier detection by exploiting Lemma~\ref{prop1}.

\vspace{-1mm}
\begin{proof}
A direct geometrical proof can be derived by considering the degrees of freedom of points in a sequence of stereo frames (see Fig.~\ref{fig:proof}). Given an arbitrary point $x^l_i$ seen by the left camera $C^l_i$, the location of its counterpart $x^r_i$ in $C^r_i$ is determined by an epipolar constraint with one DoF. The same is true for the match between $x^l_{i+1}$ and $x^r_{i+1}$. Finally the associations along time, $x^l_i \leftrightarrow x^l_{i+1}$ and $x^r_i \leftrightarrow x^r_{i+1}$, given by the motion transformation $M^i$ are determined by two more DoF, summing up to six DoF.
\end{proof}

\vspace{-3mm}
\subsection{Robust Decomposition with Constrained Rank}

We use the rank constraint in (\ref{eq:rank2}) to detect potential outliers present in the data matrix $W$. The corrupted matrix $W$ can me decomposed as a sum of two terms $W = L + S$. In this decomposition we assume that $L$ is the uncorrupted version of $W$, which satisfies the correct rank constraint. On the other hand, $S$ is a sparse matrix accounting for the outliers, i.e., ``infrequent events'' of unbounded magnitude. This decomposition can be achieved by imposing sparsity constraints on the error matrix $S$ as 

\begin{eqnarray}
\label{eq:np}
& \text{min}_{L, S} \text{ rank}(L) +  \lambda \norm{S}_{\ell_0} &\\
& \textbf{s.t. } \text{rank}(L) \leq r, \norm{W - L - S}_F \leq \epsilon. &\nonumber
\end{eqnarray}

However, this formulation becomes intractable due to the presence of the $\ell_0$-norm, whose minimization is of combinatorial nature and leads to a NP-hard optimization problem~\cite{AugmentedLagrangian}\cite{Wright09}\cite{Candes11}. Recent discoveries have led to the proposal of convex counterparts of (\ref{eq:np}), referred in the literature to as Robust PCA or Robust Principal Component Pursuit\cite{RPCAcomparison}. These techniques are based on a convex relaxation of (\ref{eq:np}) as

\vspace{-2mm}
\begin{equation}
\label{eq:rpca}
\text{min}_{L, S} \norm{L}_* +  \lambda \norm{S}_{\ell_1} \text{s.t. } W = L + S,\\
\end{equation}

\noindent where $\norm{\cdot}_*$ is the nuclear norm that acts as the convex envelope of the rank function, and the $\ell_1$-norm acts as the convex envelope of the $\ell_0$ norm. 

The convex problem (\ref{eq:rpca}), although very convenient, suffers from two conceptual limitations when applied to low-dimensional ambient spaces. First, it does not impose the required rank constraint, i.e., $\text{rank}(L) \leq r$. This could produce low-rank solutions with different ranks, generally smaller, than the one arising in our problem, and thus leading to wrong solutions. Second, and more important, (\ref{eq:np}) and (\ref{eq:rpca}) are not equivalent in those cases where the rank is close to the dimensions of the matrix $W$~\cite{Wright09}\cite{Candes11}, as in this case. The behaviour of these limits is abrupt, which is why this phenomenon is known as phase transition, having an universal character, as stated in~\cite{Universality}. For most methods, the region of applicability is approximately of the form 
\begin{align}
\rho_r &= \frac{rank(L)}{min(m,n)} \nonumber\\
\rho_s &= \frac{\norm{S}_0}{mn} \\
\textbf{s.t. } & \rho_r + \rho_s \leq k, \nonumber
\end{align}

\noindent where, $\rho_r$ is the rank fraction and $\rho_s$ is the proportion of sparsity (i.e., outliers). The constant $k$ depends on the specific formulation but it is usually  $k \approx 0.4$ (see~\cite{Wright09}\cite{Candes11}\cite{APG}). Also, RPCA convergence theorems assume that the support of $S$ is uniformly distributed, meaning that cases where the sparse components are localized in blocks, rows or columns are out of the applicability of RPCA. Nonetheless, some experiments have been reported in~\cite{RPCAOP}\cite{SOLO12}, where $S$ has support on columns or rows and is still applicable, as far as the other requirements are met. To this end,~\cite{RPCAOP} proposes the $\ell_{2,1}$-norm as a natural way to deal with outliers as columns. However, in our tests, when the ambient dimension is low, the behaviour of $\ell_{2,1}$ methods is equal or worse than the $\ell_1$ counterpart, and computationally more costly. 

Our proposal to deal with these problems consists of extending the region of applicability of $k$ by  modifying the program (\ref{eq:rpca}) to explicitly incorporate the rank constraints. To this end, we reformulate (\ref{eq:rpca}) as  

\vspace{-2mm}
\begin{equation}
\begin{aligned}
& \text{min}_{L, S} \frac{1}{2}\norm{W - L - S}_F^2 +  \lambda \norm{S}_{\ell_1} &\\
 & \textbf{s.t. } \text{rank}(L) \leq r &.
 \end{aligned}
 \label{eq:ours}
\end{equation} 

We refer to (\ref{eq:ours}) as Robust Decomposition with Constrained Rank (RD-CR). This formulation enables solving problems in harder conditions, i.e., higher ranks and greater proportions of outliers. However, in motion estimation problems, the rank is still too high to achieve an exact estimation of $L$ and $S$. For this reason, we tackle the problem by using the residual matrix $S$ to infer which columns (point matches) are outliers. The steps required to perform this process are explained in detail in section~\ref{sec:approach}.


%% file: sections/Approach.tex
\section{Motion Estimation via RD-CR}
\label{sec:approach}

This section explains our proposal of two stages framework for robust estimation of motion models (see Fig~\ref{fig:pipeline}). The first stage uses the RC-RD method as a preprocessing step and the second stage applies a compressed regression procedure that exploits data reduction to achieve high performance. 

\subsection{RD-CR as a preprocessing stage}
\label{sec:app1}

The process of outlier detection starts by solving the program (\ref{eq:ours}) with the proposed RD-CR approach, which is based on proximal gradients, following the philosophy of APG~\cite{APG}. However, in our case the use of the nuclear norm is avoided and the SVD is changed for a more convenient skinny SVD that serves to generate feasible candidates of rank $r = 6$ with a computational complexity of $O(rn^2)$ per iteration~\cite{LADM}. Our technique also serves to deal with matrices presenting a rank below $6$, as the skinny SVD will produce singular values very close to zero in those cases. This method, summarized in Algorithm~\ref{alg:ours}, consisting of three conceptual steps; namely, initialization, projected gradient descent and continuation.

\subsubsection{\textbf{Initialization} (lines 1--3)}

The imposed constrained rank makes (\ref{eq:ours}) non-convex, for which a proper initialization of $L_0$ and $S_0$ is required. A good initialization is achieved by solving the convex surrogate~(\ref{eq:rpca}) through the use of the APG method~\cite{APG}. APG uses a standard proximal algorithm improved by Nesterov acceleration~\cite{Nesterov1} and a continuation technique to speed up the convergence of the terms involving proximity operators~\cite{Continuation1}, i.e. both the nuclear and $\ell_1$ norms. Despite its convexity, it should be noticed that in the extreme cases we are addressing (high $\rho_r$) APG might not converge to a suitable solution. However, when APG is run for a fixed amount of iterations ($k=20$), it produces candidates with enough quality to initialize our method. The remaining parameters are set to $\lambda = 10^{-2}$, $\bar{\mu} = 10^{-9}$ and $k_{\text{max}} = 20$.

\subsubsection{\textbf{Projected Gradient Descent} (lines 6--13)}

The main loop of Algorithm~\ref{alg:ours} performs $k_{\text{max}}$ iterations, each one consisting of the computation of a projected descent step for $L_k$ and $S_k$. To this end, descent steps $G_k^L$ and $G_k^S$ are computed from  the Euclidean gradient $D_k = L_k + S_k - W$. $G_k^L$ is projected to the rank-6 manifold through the skinny SVD, while $G_k^S$ is projected to promote sparsity through the soft-thresholding operator 

\begin{equation}
\mathcal{S}_{\mu}[M] = \text{max}(0, M-\mu) + \text{min}(0, M+\mu).
\end{equation}

We use fixed-length steps $\alpha_L$, $\alpha_S$ for the descent process with values in the ranges $\alpha_L \in [1/10, 1]$, $\alpha_S \in [1/10, 1/2]$, which are close to the Lipschitz constant of (\ref{eq:ours}).

\subsubsection{\textbf{Continuation} (lines 4 and 14--16)}
Continuation is defined as a technique to improve the convergence of proximity operators like the soft-thresholding by dynamically adapting the parameter $\mu_k$ following a geometrical series~\cite{Continuation1}\cite{APG}. However, in our approach we adapt the continuation scheme to be more coherent with the current estimation. This is done by taking $\mu_k$ as a fraction of the mean size of the singular values of the Euclidean gradient $D_k$, as, 

\begin{equation}
\mu_D = \delta \norm{D_k}_F / \sqrt{mn}, \quad \delta = 10^{-3}.
\end{equation}

Then, $\mu_D$ is used to generate a new $\mu_{k+1} = \max(\frac{1}{\lambda}\mu_D,\bar{\mu})$, which empirically led to a higher performance.

\vspace{-0mm} 
\begin{algorithm}
 \algsetup{linenosize=\small}
  \small
\begin{algorithmic}[1]
        \REQUIRE Data matrix $W\in \mathbb{R}^{m\times n}$, initialization $(L^{\rm APG},S^{\rm APG})$, $\lambda$, $k_\text{max}$, $\delta$, $\bar{\mu}$.
	\STATE $k \leftarrow 0$
	\STATE $L_k \leftarrow L^{\rm APG}$, $S_k \leftarrow S^{\rm APG}$, $\alpha_L \leftarrow 1$, $\alpha_S \leftarrow 1 / 5$
	\STATE $(U_0, \Sigma_0, V_0) \leftarrow svds(W, 6)$, $\tilde{W} \leftarrow U_0 \Sigma_0 V_0^T$
	 \STATE $\mu_k \leftarrow \delta \norm{W-\tilde{W}}_F / \sqrt{mn}$.
	\WHILE{$k < k_{\text{max}}$}
 	\STATE $D_k \leftarrow L_k + S_k - W$
 	\STATE \textit{\color{light-gray}//Update of L as a descent gradient step + projection} 	
	\STATE $G_k^L\leftarrow L_k -\alpha_L D_k$.
	\STATE $(U_{k+1},\Sigma_{k+1},V_{k+1})\leftarrow svds(G_k^L,6)$
	\STATE $L_{k+1}=U_{k+1} \Sigma_{k+1} V_{k+1}^T$.
	\STATE \textit{\color{light-gray}//Update of S as a descent gradient step + projection}
        \STATE $G_k^S\leftarrow S_k - \alpha_S D_k$.
         \STATE $S_{k+1} \leftarrow \mathcal{S}_{\mu_k}[G_k^S]$.
       \STATE \textit{\color{light-gray}//Update of } $\mu_k$ \textit{\color{light-gray} with continuation}
       \STATE $\mu_D \leftarrow \norm{D_k}_F / \sqrt{mn}$
          \STATE $\mu_{k+1} \leftarrow \max(\frac{1}{\lambda}\mu_D,\bar{\mu})$.
         \STATE $k \leftarrow k+1$.
	\ENDWHILE
	\RETURN{$L\leftarrow L_k$, $S\leftarrow S_k$. }
\end{algorithmic}
\caption{RD-CR with Proximal Gradients}
\label{alg:ours}
\end{algorithm}

A proof of convergence for the Algorithm~\ref{alg:ours} is very hard due to the large value of $\rho_r$ for our low ambient dimension scenario. However, this does not limit its applicability to real problems  since in our experiments running the algorithm during $k_\text{max} = 20$ iterations has produced good results. Moreover, the proposed method is very efficient due to the moderate computational complexity of Algorithm~\ref{alg:ours}, i.e. $O(rn^2)$, and given that $r=6$ and $n=8$. 

After applying Algorithm~\ref{alg:ours}, the resultant $S$ contains outliers distributed in columns. Then, in order to decide if the $j$-column (point match) is an outlier or not, we apply the following decision criterion,

\begin{equation}
\label{eq:decision}
\psi_S(x_j) : \mathbb{R}^8 \rightarrow \{0,1\}
\end{equation}

\begin{equation}
\label{eq:decision2}
 \psi_S(x) =
  \begin{cases} 
      \hfill 1,    \hfill & \text{ if } \norm{x}_{\ell_1} > \text{min}(\tau_0, \frac{1}{N_c} \norm{S}_{\ell_1}) \\
      \hfill 0, \hfill & \text{otherwise}. \\
  \end{cases}
\end{equation}

The decision made by $\psi_S(\cdot)$ is based on a threshold directly proportional to the average $\ell_1$-norm of the matrix $S$ up to a saturation point $\tau_0 = 0.5$. This expression serves to adapt the threshold to the mean residual of the data, what in our experiments has reported the best performance through cross-validation. In the cases where $\psi_S(x_j) = 1$, the $j-th$ match point is removed from the correspondence set.

\begin{figure*}[!b]
\normalsize
\newcounter{MYtempeqncnt}
\setcounter{MYtempeqncnt}{\value{equation}}
\setcounter{equation}{23}
\begin{equation}
\label{eq:w}
	A^{m,(j)} =   \scalemath{0.9}{\begin{bmatrix}
[0]_{1 \times 3} & f {X_{i}^{l,(j)}}^T & (c_v - v_{i+1}^{m,(j)}) {X_{i}^{l,(j)}}^T & 0 & f & (c_v - v_{i+1}^{m,(j)}) & 0\\
	-f {X_{i}^{l,(j)}}^T & [0]_{1 \times 3} & (u_{i+1}^{m,(j)} - c_u) {X_{i}^{l,(j)}}^T & -f & 0 & (u_{i+1}^{m,(j)} - c_u) & \alpha\\
	f v_{i+1}^{m,(j)} {X_{i}^{l,(j)}}^T & -f u_{i+1}^{m,(j)} {X_{i}^{l,(j)}}^T & (c_u v_{i+1}^{m,(j)} - c_v u_{i+1}^{m,(j)}) {X_{i}^{l,(j)}}^T & f v_{i+1}^{m,(j)} & -f u_{i+1}^{m,(j)} & (c_u v^{(j)} - c_v u_{i+1}^{m,(j)}) & \beta\\
	\end{bmatrix}}
\end{equation}
\setcounter{equation}{\value{MYtempeqncnt}}
\hrulefill
\vspace*{4pt}
\end{figure*}

\subsection{Fast Compressed Least-Squares}
\label{sec:app2}

The second stage of our framework consists in the actual estimation of the transformation $M_i$. In order to make this stage as fast as possible we consider those cost functions that are easy to linearise with respect to the unknowns. We have selected the algebraical cost function proposed in~\cite{Ros2013b}, defined as;

\vspace{-6mm}
\begin{align}
\label{eq:cost1}
 E(M) =& \sum_{j=1}^{N} \norm{  \left( \Pi \text{K } M \hat{X}^{l,(j)}_{i} \right) \times \hat{x}^{l,(j)}_{i+1} }^2 +\\ 
   &\norm{ \left( \Pi \text{K } M \hat{X}^{l,(j)}_{i} -{\vec{B}} \right) \times \hat{x}^{r,(j)}_{i+1} }^2. \nonumber
\end{align}

This function measures the reprojection error of points in the two stereo views as the misalignment of homogeneous vectors. Since this cost is algebraical instead of a geometric one, it requires a proper normalization of the data, i.e., removing its mean and scaling it up by its standard deviation as described in~\cite{AlgError}\cite{MultipleView}. In this context, $\hat{X}$ stands for homogeneous 3D points and $\hat{x}$ for homogeneous 2D points. $\times$ is the standard cross product of vectors. The rest of the notation remains as explained before. 

The choice of (\ref{eq:cost1}) is motivated by its tolerance to be efficiently compressed. Compression consists in turning (\ref{eq:cost1}), which sums over all the observations, into a compact equivalent $\Gamma$, known as  Reduced Measurement Matrix (RMM). In this case, the compression technique used is fully equivalent to the original data, i.e., is a lossless compression. The compression step is performed by following the process,
\begin{multline}
\label{eq:rmm}
E(M)  = \sum_{j=1}^{N} \norm{  \left( \Pi \text{K } M \hat{X}^{l,(j)}_{i} \right) \times \hat{x}^{l,(j)}_{i+1} }_{\ell_2}^2 + \\
   \norm{ \left( \Pi \text{K } M \hat{X}^{l,(j)}_{i} -{\vec{B}} \right) \times \hat{x}^{r,(j)}_{i+1} }_{\ell_2}^2 = \\
  \sum_{j=1}^{N} \norm{A^{l,(j)} \breve{M}}_{\ell_2}^2 +  \norm{A^{r,(j)} \breve{M}}_{\ell_2}^2 =\\ 
   \norm{\boldsymbol{A^l} \breve{M}}_{\ell_2}^2 + \norm{\boldsymbol{A^r} \breve{M}}_{\ell_2}^2 = \\
   \breve{M}^T \underbrace{\boldsymbol{A^l}^T \boldsymbol{A^l}}_{\boldsymbol{\Gamma^l}} \breve{M} + \breve{M}^T \underbrace{\boldsymbol{A^r}^T \boldsymbol{A^r}}_{\boldsymbol{\Gamma^r}} \breve{M} = \breve{M}^T \boldsymbol{\Gamma} \breve{M}.
\end{multline}

For our particular problem $\Gamma$ is a $13 \times 13$ matrix, which is determined from the number of coefficients required to linearise the original cost function (\ref{eq:cost1}). Here, $A^{m,(j)}$, for $m={l,r}$ is the matrix of linearised coefficients which results from combining the calibration $K$, $\Pi$ and the observations, as shown in (\ref{eq:w}). Then $\boldsymbol{A^m}$ is the stack of the $N$ matrices $A^{m,(j)}$ and $\breve{M} = \left[\textbf{stack}(R), T, 1\right]^T$, i.e., a vector resulting from stacking the components of $M$ with an homogeneous component.

The reduction of (\ref{eq:rmm}) is very efficient due to its algebraical nature and just has to be performed once, before the optimization takes place. Then, the optimization is posed as a compressed least-squares problem on the $\mathbb{SE}(3)$ manifold in order to fulfil the orthonormal constraint of 3D rigid motions. This can be formulated  as,

\vspace{-3mm}
\begin{align}
\label{eq:lm}
\omega^* &= \argmin_\omega E(M) = \argmin_\omega  \breve{M}^T \boldsymbol{\Gamma} \breve{M} \\
&= \argmin_\omega  \textbf{stack}[\text{exp}_{\mathbb{SE}(3)}(\omega)]^T \ \boldsymbol{\Gamma} \  \textbf{stack}[\text{exp}_{\mathbb{SE}(3)}(\omega)], \nonumber
\end{align}

\noindent where $\omega^*$ is the estimated motion transformation given as an element of the Lie algebra, $\omega \in \mathfrak{se}(3)$, which is the tangent space of $\mathbb{SE}(3)$ at the identity. The use of the exponential map serves to move from the tangent space back to the manifold. This map is given by the Rodrigues formula~\cite{ros13} and is a requirement to perform the evaluation of the cost function. Despite the special properties of (\ref{eq:lm}), it can be easily optimized by using Levenberg-Marquardt. The only requirement for this optimization is the computation of the Jacobian of (\ref{eq:lm}), something that is done by computing its partial derivatives as,

\begin{align}
\label{eq:part}
\partial_{\omega_i} E(M) \partial_{\omega_i} \xi^T (\boldsymbol{\Gamma} + \boldsymbol{\Gamma}^T) \xi,\\ \text{where } \xi = \textbf{stack}[\text{exp}_{\mathbb{SE}(3)}(\omega)] = \breve{M}.\nonumber
\end{align}

The partial derivatives of $\text{exp}_{\mathbb{SE}(3)}$ are calculated by considering a $k$-order approximation (with $k \approx 10$ in our experiments) of the expression from the Taylor's series defined as,

\begin{align}
\label{eq:jacobian}
\partial_{\omega_i} \mathbf{exp}(Z) \approx \partial_{\omega_i} \left(\sum_{n \geq 0}^k \frac{Z^n}{n!}\right) =\notag\\
\sum_{n \geq 1}^k{\frac{1}{n!}\left(  \partial_{\omega_i} Z \ Z^{n-1} + Z \ \partial_{\omega_i} Z^{n-1} \right)}\\
\text{with } Z = \begin{bmatrix} 0 & -\omega_3 & -\omega_2 & \omega_4\\ \omega_3 & 0 & -\omega_1 & \omega_5\\ -\omega_2 & \omega_1 & 0 & \omega_6\\ 0 & 0 & 0 & 1 \end{bmatrix}.
\end{align}

The optimization of this function is very fast and converges in a couple of iterations. The use of $\Gamma$ reduces each iteration to just various matrix by vector products and  vector by vector products, what makes the compressed least-squares procedure very efficient. The total time for the compression and the optimization is around one millisecond in a standard desktop computer. 

\addtocounter{equation}{1}

%% file: sections/Results.tex
\section{Experimental Evaluation}
\label{sec:results}

This section shows the experiments performed on synthetic and real data to validate our approach. On one hand, synthetic data is used as a control environment to analyse the behaviour of the RD-CR algorithm in different situations. On the other hand, experiments with real data serve to compare the performance of the whole framework against state-of-the-art techniques for stereo Visual Odometry. From the results we conclude that our approach is competitive against state-of-the-art methods in terms of accuracy and is more efficient in terms of computation.

\subsection{Synthetic data tests}

\begin{figure*}[!t]
	\centering
	\vspace{-0mm}
	\includegraphics[scale=1.27]{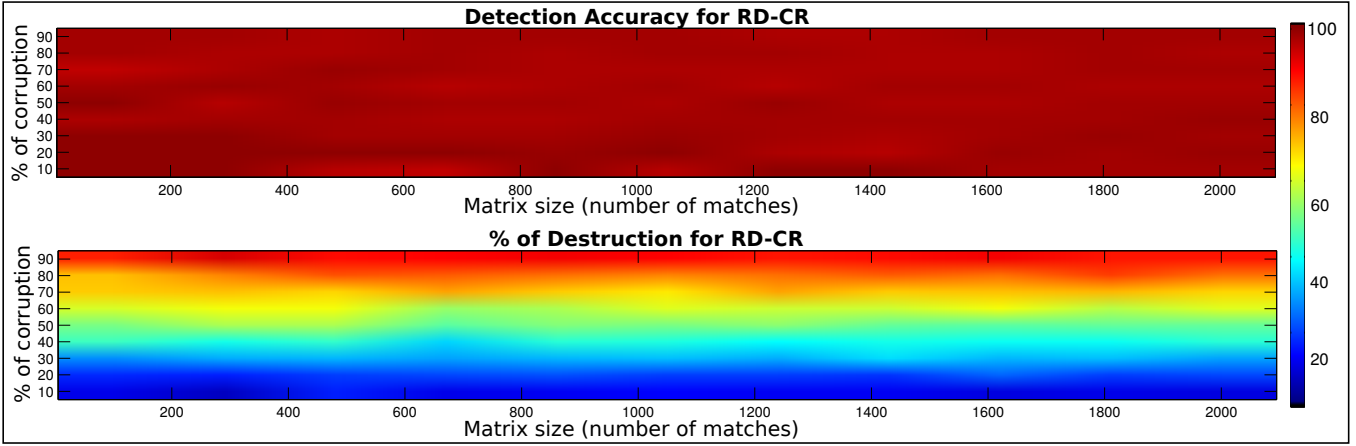}
	\vspace{-2mm}
	\caption{\textbf{Top}, mean accuracy evaluation of RD-CR w.r.t. the number of matches and the percentage of outliers. \textbf{Bottom}, mean percentage of elimination made by RD-CR for each case.}
	\label{fig:phase}
\end{figure*}

\begin{figure*}[]
	\centering
	\vspace{-0mm}
	\includegraphics[scale=1.30]{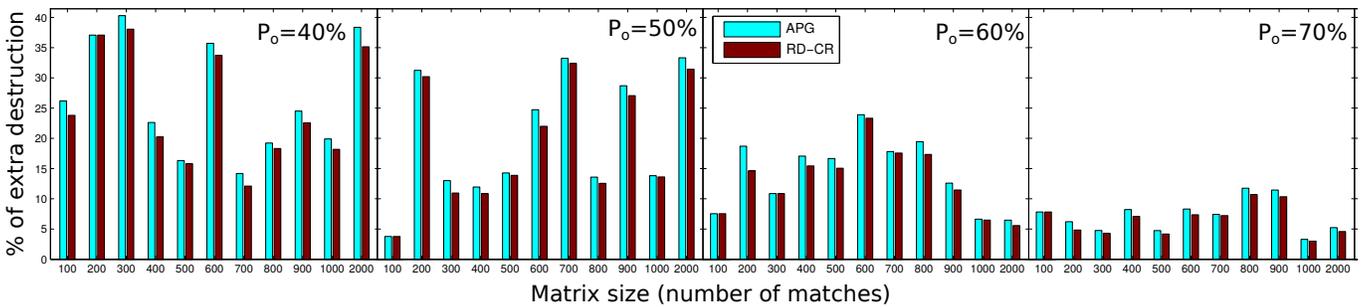}
	\vspace{-2mm}
	\caption{Comparison of APG and RD-CR regarding the percentage of excessive elimination of outliers.}
	\label{fig:versus}
\end{figure*}

We have created a set of tests that simulate the conditions of a stereo rig performing a rigid motion in 3D, which is the problem of application addressed in this paper. For each case, we evaluate the method for a specific matrix size $N_c$ (i.e., number of matches) and proportion of outliers $P_o = \rho_s$. We define the stereo rig with the reference camera matrices $(C^l_i, C^r_i)$. This means that $C^l_i$ is the reference frame, while $C^r_i$ is set at baseline distance $B$. Then, $N_m > 50$ motion transformations $M_i$ are created such that each $M_i \in \mathbb{SE}(3)$. For  each model $M_i$ we create $N_c$ matches $\{x_i^l, x_i^r, x_{i+1}^l, x_{i+1}^r\}$, according to the four views. Point matches are then contaminated with random noise of magnitude $\sigma_n = 1.5$ and impulsive noise with magnitude $\sigma_j \in [2, 100]$, to represent outliers. In each case the $P_o\%$ of elements $x_i^l, x_i^r, x_{i+1}^l, x_{i+1}^r$ are corrupted by $\sigma_j$ in all their components. This data is reorganized to form a matrix $W_i \in \mathbb{R}^{8 \times N_c}$ as explained in Sec.~\ref{sec:problem}.

We have run this test repeatedly for values of $N_c$ from $100$ to $2000$ matches and ratios of outliers $P_o$ from $10\%$ to $90\%$.  In each combination of $N_c$ and $P_o$, $50$ repetitions are performed and from them the average of outlier detection is calculated, i.e., the accuracy of the method. The results of this experiment are shown in Fig.~\ref{fig:phase}\textbf{(Top)}. Our approach RD-CR shows a high success ratio, identified by warm colours, for the different sizes of matches considered and percentages of outliers. Additionally, as we can see in  Fig.~\ref{fig:phase}\textbf{(Bottom)}, the percentage of outlier removal of RD-CR closely matches the expected percentage of corruption, meaning that the method recall is high and it does not produce many false positives. 

By imposing the correct rank, RD-CR reduces the percentage of excessive elimination of outliers, a problem that is present in APG, as shown in Fig.~\ref{fig:versus}. Given several levels of corruption we compare the excess of outlier removal (false positives) caused by RD-CR and APG. From these results we can conclude that, in general, RD-CR overcomes the efficacy of APG to detect outliers and causes less false positives for the different configurations of noise. This is mainly due to the correct rank estimation done by RD-CR, as APG tends to underestimate the rank, leading to more false positives.

This synthetic experiment aims to show that it is possible to exploit RPCA-like methods to detect outliers even when working beyond the recovery boundaries of these techniques, which is the initial assumption of this work. In these cases, the high accuracy of our approach at outlier detection is closely linked to a precise selection of the threshold $\tau_0$ within the function $\psi_S$. However, when working in real scenarios this information cannot be determined in a precise fashion. In addition, real scenarios bring extra challenges, such as multiple corruption per correspondence, i.e., more than one point is severely affected within a match, and $\sigma_j$ can adopt moderated values, making the detection process more challenging. Nevertheless, our next experiments prove that RD-CR still work in real scenarios under these adverse conditions.


\subsection{Real data tests}

We now show the good performance of our framework applied to stereo Visual Odometry in real sequences. Data sets are taken from the challenging KITTI benchmark~\cite{KITTI} in its Visual Odometry modality. KITTI sequences correspond to large urban environments, providing stereo images taken from a car. We use  sequences 00--10 for which ground truth is available. 

The comparison is carried out with 5 state-of-the-art methods for motion estimation:  RANSAC~\cite{Ransac} and L1-averaging~\cite{L1rotation} representing classical consensus and modern consensus-by-averaging methods. IRLS-L1~\cite{Leary90} and Huber M-estimator~\cite{Huber} as representatives of fast robust estimators, and Compressed Least-squares (C-LS) ---the compressed flavour of least-squares presented in section~\ref{sec:app2}--- stands for non-robust methods. In addition, APG is also tested in order to highlight the difference in accuracy when the rank constraint is not imposed. Other methods, like Bundle Adjustment, are excluded since our goal is the estimation of individual models, without considering the temporal relationships among the different motion transformations. 

In our experiments RANSAC and L1-averaging are set to produce a maximum of $250$ and $1000$ models respectively during the consensus, which is a realistic number for real-time applications. The constant for the Huber M-estimator is set to $\delta = 2.5$, after a process of validation. The number of point matches per frame is around $2000$. For all methods the computation of the models is done by an optimization process on $\mathbb{SE}(3)$, via a couple of iterations of Levenberg--Marquardt. The cost function used in this optimization follows a standard projection of 3D points into the left and right cameras~\cite{VisualOdometryStereo}, over a fixed number of matches $N_k = 3$, i.e., the minimum amount required to generate a valid model. In our experiments this cost function is defined as,

\vspace{-4mm}
\begin{align}
\argmin_\omega \sum_{i=1}^{N_k} \norm{  \Pi \text{K } \text{exp}(\omega) {X}^{l,(j)}_{i} - {x}^{l,(j)}_{i+1} }_{\ell_2}^2 + \\
\norm{ \Pi \left( \text{K } \text{exp}(\omega) {X}^{l,(j)}_{i} -{\vec{B}} \right) - {x}^{r,(j)}_{i+1} }_{\ell_2}^2. \nonumber
\end{align}

As an error metric we use the mean of the individual relative errors along a whole sequence. To measure this error, we have considered the most natural metric, i.e., the group structure of rigid transformations, which is defined as;

\begin{equation}
\mathcal{E}(M_i, M^*)=\norm{log_{\mathbb{SE}(3)}(M_i M^{*^{-1}})}_{\ell_2}.
\end{equation}

In this expression, $log_{\mathbb{SE}(3)}(\cdot)$ is the logarithm of the $\mathbb{SE}(3)$ group, mapping from the group to its tangent space. This tool is used to measure the difference between the estimated motion $M_i$ and the ground truth $M^*$. From that, a relative error is computed to facilitate an intuitive interpretation,

\vspace{-4mm}
\begin{equation}
\mathcal{E}_\text{rel}(M_i, M^*)= \frac{\norm{log_{\mathbb{SE}(3)}(M_i M^{*^{-1}})}_{\ell_2}}{\norm{log_{\mathbb{SE}(3)}(M^{*})}_{\ell_2} + \epsilon}, \text{ for } \epsilon = 10^{-5}.
\end{equation}

\begin{figure*}[!t]
	\centering
	\includegraphics[scale=0.90]{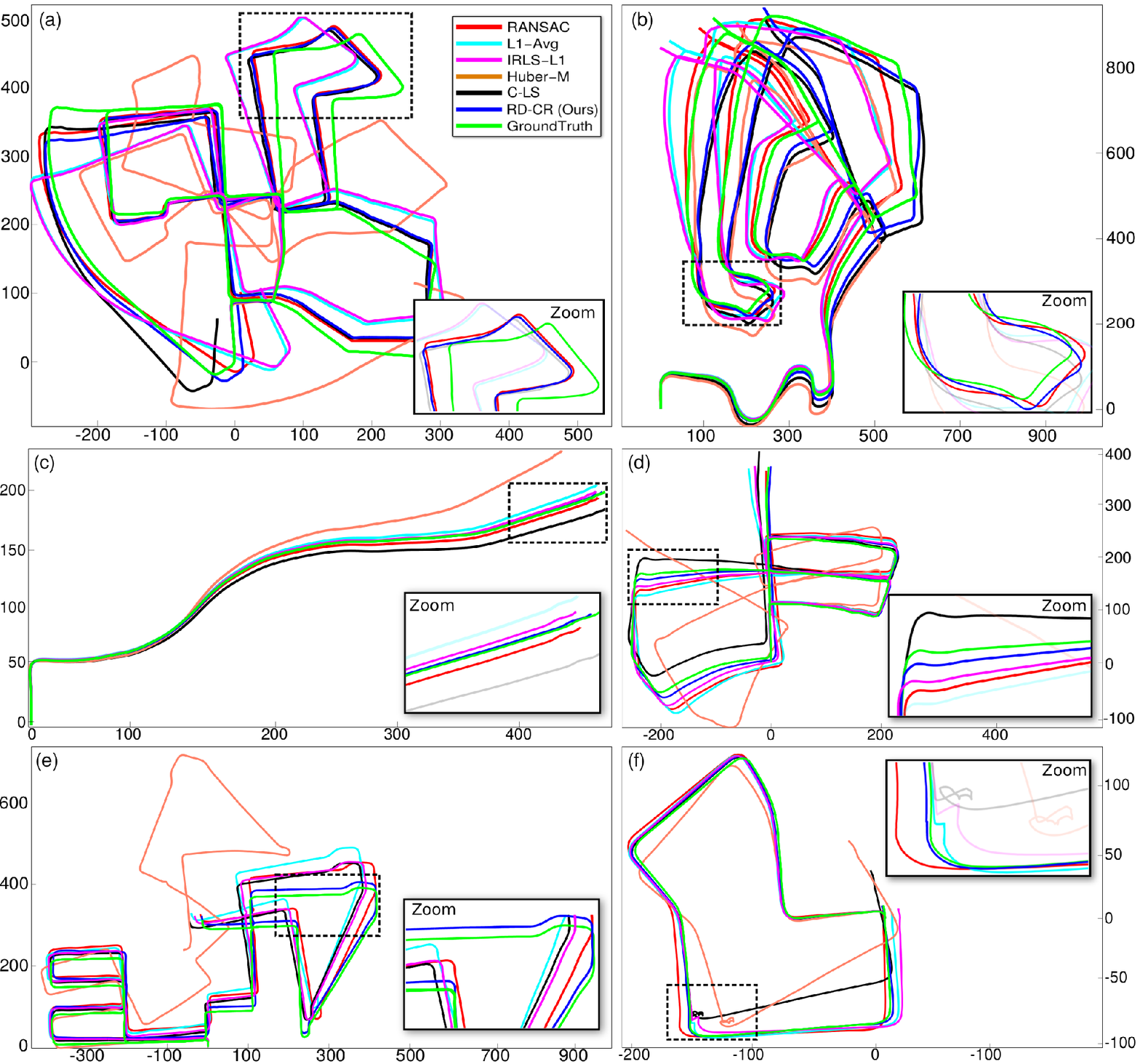}
	\caption{Visual Odometry results for sequences 00 \textbf{(a)}, 02 \textbf{(b)}, 03 \textbf{(c)}, 05 \textbf{(d)}, 08 \textbf{(e)} and 07 \textbf{(f)}.}
	\label{fig:VOResults}
\end{figure*}

\begin{figure*}[!t]
	\centering
	\includegraphics[scale=0.90]{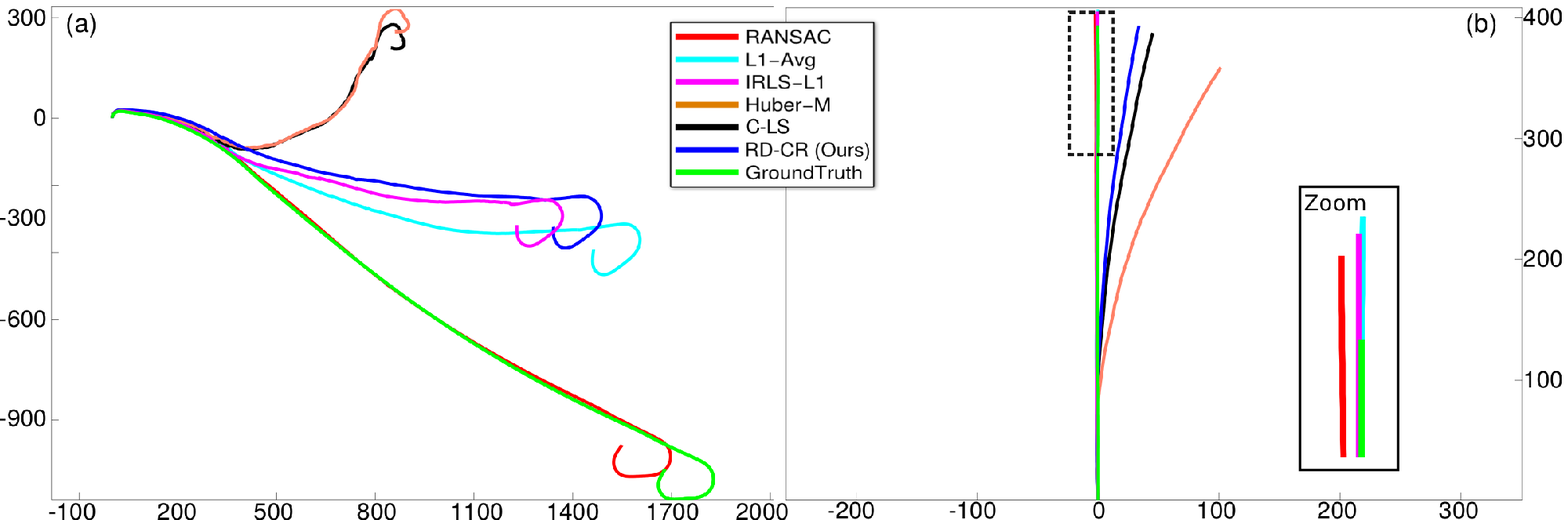}
	\caption{Visual Odometry results for sequences 01 \textbf{(a)}, and 04 \textbf{(b)}.}
	\label{fig:bad}
	\vspace{-3mm}
\end{figure*}

It should be stressed that this metric provides a measure of the local error, and does not take into account the accumulative effect of the error in the global trajectories. This point is critical in order to understand the differences between the results provided by Table~\ref{tab:kitti}, which is focused on local errors; and Fig.~\ref{fig:VOResults} and Fig.~~\ref{fig:bad}, which show the accumulative effects of the errors. 

\renewcommand*\arraystretch{1}
\begin{table*}[t]
\vspace{-0mm}
\begin{center}
\caption {Evaluation of methods measuring the mean relative error for  KITTI sequences (00-10).}
\vspace{-2mm}
\small
\tabcolsep=0.17cm
\begin{tabular}{|l|c|c|c|c|c|c|c|c|c|c|c|c|}
\hline 
 & 00 & 01 & 02 & 03 & 04 & 05 & 06 & 07 & 08 & 09 & 10 &\textbf{Global}\\
 \hline
\multicolumn{13}{|c|}{\textbf{Consensus based Methods} }\\ 
\hline 
RANSAC & 5.27\% & 10.05\% & 4.83\% & 5.97\% & 5.30\% & 8.47\% & 5.52\% & 13.89\% & 7.77\% & 5.45\% & 5.80\% & 6.71\%\\
\hline
L1-Avg. & 5.44\% & 19.55\% & 4.95\% & 6.34\% & 5.72\% & 8.23\% & 5.92\% & 13.08\% & 8.26\% & 5.84\% & 6.79\% & 7.35\%\\
\hline
\multicolumn{13}{|c|}{\textbf{Non-Consensus based Methods}}\\
\hline
IRLS-L1 & 5.35\% & 30.68\% & 4.86\% & 6.08\% & 5.44\% & 7.66\% & 5.63\% & 13.22\% & 7.94\% & 5.57\% & 7.35\% & 7.71\%\\
\hline
Huber-M. & 5.30\% & 34.40\% & 4.80\% & 6.10\% & 9.20\% & 11.89\% & 5.59\% & 25.61\% & 12.95\% & 6.64\% & 7.97\% & 9.88\%\\
\hline
C-LS & 5.77\% & 51.52\% & 5.05\% & 7.23\% & 6.45\% & 40.99\% & 6.32\% & 135.32\% & 10.70\% & 6.07\% & 8.10\% & 19.22\%\\
\hline
\multicolumn{13}{|c|}{\textbf{Robust Decomposition based Methods}}\\
\hline
APG & 10.19\% & 22.78\% & 7.59\% & 9.11\% & 7.27\% & 16.74\% & 6.67\% & 24.14\% & 13.67\% & 10.49\% & 10.64\% & 12.12\%\\ 
\hline
\textbf{RD-CR (Ours)} & 5.42\% & 27.23\% & 4.76\% & 6.02\% & 5.69\% & 7.85\% & 5.36\% & 12.15\% & 8.00\% & 5.46\% & 6.32\% & {7.49}\%\\
\hline
\end{tabular} 
\label{tab:kitti} 
\end{center}
\vspace{-4mm}
\end{table*}

\singlespacing
\begin{table*}[t]
\begin{center}
\caption {Computational time analysis for the methods under comparison. The evaluation is carried out in Matlab by measuring the different parts of each method independently and keeping the size of the loops small to provide faithful comparisons. }
\vspace{-3mm}
\tabcolsep=0.09cm
\scriptsize
\begin{tabular}{|l|c|c|c|}
\hline 
\textbf{Methods} &  \multicolumn{2}{|c|}{\textbf{Time analysis} } & \textbf{Time~(s)}\\
\hline
RANSAC & \medmuskip=0mu \thickmuskip=0mu $\mathclap{} N_{\text{model}} \times (T_\text{ModelGeneration} + N_{\text{points}} \times T_\text{CheckInliers})$ & $250 \times (1.3e{-3} + 2000 \times 5e{-5})$ & $25.32$\\
\hline
\multirow{2}{*}{L1-Avg} & \medmuskip=0mu \thickmuskip=0mu $N_{\text{model}} \times T_\text{ModelGeneration} + T_\text{InitL1} +$ & $1000 \times 1.3e{-3} + 1.7e{-2} +$ & $15.85$\\
& \medmuskip=0mu \thickmuskip=0mu $ \text{Iters} \times ( N_{\text{model}} \times T_\text{ProcessModel} + T_\text{RetractionMap})$ & $ 200 \times (1000 \times 7.26e{-5} + 4.7e{-5})$ & \\
\hline
IRLS-L1 & \medmuskip=0mu \thickmuskip=0mu $\text{Iters} \times  N_{\text{points}}  \times (T_\text{CostJacobian} + T_\text{Reweighting})$ &  $15 \times 2000 \times (1.2e{-4} + 2.8e{-5})$ & $4.44$\\
\hline
Huber-M & \medmuskip=0mu \thickmuskip=0mu $\text{Iters} \times  N_{\text{points}}  \times T_\text{CostJacobian}$ &  $15 \times 2000 \times 1.2{-4}$ & $3.60$\\
\hline
C-LS & \medmuskip=0mu \thickmuskip=0mu $T_\text{CreationRMM} + \text{Iters} \times T_\text{CostJacobian} + T_\text{Refinement}$ &  $1.6e{-2} + 5 \times 6.2e{-5} + 2.6e{-1}$ & $0.28$\\
\hline
APG & \medmuskip=0mu \thickmuskip=0mu $\text{Iters} \times (T_\text{Gradients} + T_\text{SVD-ECON} + T_\text{Nuclear} +  T_\text{Proximity} + T_\text{Nesterov}) + T_\text{C-LS}$ &  $100 \times (1.2e{-3} + 1.37e{-3} +3e{-3}+ 7.13e{-4} + 8.75e{-4}) + 0.28$ & $0.99$\\
\hline
\multirow{2}{*}{\textbf{RD-CR (Ours)}} & \medmuskip=0mu \thickmuskip=0mu $\text{ItersAPG} \times (T_\text{Gradients} + T_\text{SVD-ECON} + T_\text{Nuclear} +  T_\text{Proximity} + T_\text{Nesterov}) + $ &  $20 \times (1.2e{-3} +1.37e{-3} +3e{-3}+ 7.13e{-4} + 8.75e{-4}) +$ & $0.47$\\ 
& \medmuskip=0mu \thickmuskip=0mu $\text{ItersRDCR} \times (T_\text{Gradients} + T_\text{SSVD} + T_\text{Proximity}) + T_\text{C-LS}$ &  $20 \times (2.64e{-4} + 1.4e{-3} + 7.13e{-4}) + 0.28$  & \\
\hline
\end{tabular}%
\label{tab:times} 
\end{center}
\vspace{-3mm}
\end{table*}

\tymin=51pt
\tymax=\maxdimen
\begin{table*}[!ht]
  \centering
  \caption{Time legend, listing the different components of each algorithm.}
  \vspace{-3mm}
  \scriptsize
  \begin{tabulary}{\linewidth}{|C|L|C|L|}
    \hline
 $N_{\text{model}}$ & Number of model generated from random matches. & $T_\text{ModelGeneration}$ & Time used to generate each model by means of an iterative process on $\mathbb{SE}(3)$.\\
 \hline
 $N_{\text{points}}$ & The amount of point matches considered. & $T_\text{CheckInliers}$ & Time employed to check the membership of a specific match to the inlier set.\\
  \hline
 $T_\text{InitL1}$ & Time taken by the initialization of the $\ell_1$-averaging. &  $\textit{Iters}$ & Average number of iterations of a given method\\
 \hline
 $T_\text{ProcessModel}$ & Time for the projection of a model to the tangent space $\mathfrak{se}(3)$. & $T_\text{RetractionMap}$ & Time needed to map a model from its tangent space to the $\mathbb{SE}(3)$ manifold.\\
  \hline
 $T_\text{CostJacobian}$ & Time used to compute a cost function and its partial derivatives. & $T_\text{Reweighting}$ & Time used for updating the weights of $IRLS-L1$ after each iteration.\\
 \hline
$T_\text{CreationRMM}$ & Time for compressing the initial matches into the Reduced Measurement Matrix. & $T_\text{Refinement}$ & Time used to perform an optional refinement based on the re-estimation of the inlier set.\\
 \hline
$\textit{ItersAPG}$ & Number of iterations needed by APG. & $\textit{ItersRDCR}$ & Number of iterations needed by RD-CR.\\
 \hline
$T_\text{SVD-ECON}$ & Time to compute an economic SVD. & $T_\text{SSVD}$ & Time to compute a skinny SVD (compact SVD plus thresholding for $\text{rank}=6$).\\
 \hline
$T_\text{Gradients}$ & Time needed to compute the Riemannian gradients. & $T_\text{Nuclear}$ & Time to compute the nuclear norm.\\
 \hline
$T_\text{Proximity}$ & Time to compute the soft-thresholding operator. & $T_\text{Nesterov}$ & Time used to perform the computation of approximate points by following Nesterov's rule.\\
 \hline
$T_\text{C-LS}$ & Time required to perform the Compressed Least-squares method. & &\\
\hline
  \end{tabulary}
  \vspace{-4mm}
\end{table*}

\vspace{-3mm}
Table~\ref{tab:kitti} summarizes the average results of the selected methods when applied to the KITTI sequences for the task of Visual Odometry. The analysis of this table reveals that RD-CR leads to results at the level of the consensus methods. Moreover, our proposal is more efficient since it does not require to generate multiple hypotheses, being able to detect outliers from the structure of the data and leading to a promising trade-off (see Table~\ref{tab:times}). 

As shown, RD-CR produces better results than the L1-averaging for all the sequences but seq. 01. RD-CR also performs similar to RANSAC, achieving better results for sequences 02, 05 and 07. In general, our approach outperforms robust regression methods like IRLS-L1 and the Huber M-estimator. In challenging scenes, such as  sequence 01, 07 or 10, RD-CR produces significantly better results, while the IRLS-L1, Huber method and C-LS are seriously affected by outliers.

Our results also show that skipping the rank constraint has a dramatic negative influence in the outlier detection process. For this reason, APG performs dramatically worse than the proposed RD-CR technique. There are two phenomena involved in this problem. When the  rank is overestimated, APG will miss outliers (false negatives), since they will be absorbed into the low-rank matrix $L$. On the contrary, if APG underestimates the rank, the number of linear dependencies in $L$ will increase and some elements will migrate to $S$ (false positives), losing some inliers needed for the good estimation of the solution.


Figure~\ref{fig:VOResults} shows results from the point of view of the estimated trajectories. The analysis of the first two plots, Fig.~\ref{fig:VOResults}\textbf{(a)}--\textbf{(b)} shows that the accuracy of our proposal is at the level of RANSAC or even higher. This is depicted in the zoomed region of Fig.~\ref{fig:VOResults}\textbf{(b)}, where the trajectory estimated by RD-CR is closer to the ground truth. This behaviour is repeated for the trajectories shown in Fig.~\ref{fig:VOResults}\textbf{(c)},\textbf{(d)},\textbf{(e)} and \textbf{(f)}, where the difference of RD-CR with respect to the others is more pronounced. In the last sequences \textbf{(c)}--\textbf{(f)}, the trajectories estimated by RD-CR are several meters closer to the ground truth than those from the state-of-the-art methods.

Finally, we would like to consider the failure cases. In the case of seq. 01, depicted in Fig.~\ref{fig:bad}\textbf{(a)}, we attribute the bad results to the severe visual aliasing found in this highway scene, a phenomenon leading to a very high ratio of matches that are outliers but behave similarly to inliers, i.e., they are mistaken for a valid $\mathbb{SE}(3)$ motion. It is also fair to highlight that robust estimator like IRLS-L1 perform very well in some sequences. For instance, in seq. 04 (see  Fig.~\ref{fig:bad}\textbf{(b)}), IRLS-L1 produces excellent results. The main problem RD-CR finds in this sequence is the matches corresponding to a car moving at the same speed that our vehicle. This subtle motion is wrongly combined with the predominant motion transformation leading to a small but continuous drift through the sequence. We are currently working on a way to solve this specific issue for future versions of the approach.

\subsubsection{Complexity evaluation}

We present the excellent computational performance of the proposed approach by comparing it against selected state-of-the-art techniques. A special effort has been made in order to produce a fair comparison of the methods. Since all the algorithms are coded in Matlab one needs to account for the bad scalability of loops, which usually lead to a super-linear cost with respect to the loop length. To avoid this problems we analyse iterative blocks in terms of their average number of iterations and the average cost of each iteration. The average cost per iteration is calculated by  using short loops, to avoid the region of bad scalability. 

The results of our analysis are summarized in Table~\ref{tab:times}. For each method we highlight the most important blocks that are involved in the computational performance of the approaches. 

In this evaluation, RD-CR shows to be one of the most efficient approaches, just behind the non-robust C-LS, which is the second stage of RD-CR. Consensus models spend a large amount of time at generating putative models and checking the pertinence of matches to the inlier set. This is specially noticeable in our experiments since the number of matches is large ($N_{\text{points}} \approx 2000$). Such a choice is justified by works like~\cite{Ros2013b}, showing that managing large amounts of matches improves stability and accuracy (up to a limit). Methods based on least-squares, such as the IRLS and the Huber M-estimator suffer from a similar problem due to the computation of large Jacobians. 

The behaviour of APG and RD-CR is quite different to the previous methods. These approaches do not deal with individual instances of the data, but with the whole data set as a single entity. This conception allows APG and RD-CR to be very efficient even when having to face a large number of matches. Furthermore, due to the structure of RD-CR, the number of iterations required to its convergence is less than the iterations needed by APG; that is, RD-CR achieves a higher accuracy with less number of iterations, being computationally very appealing. 

%% file: sections/Conclusions.tex
\section{Conclusion and future work}
\label{sec:conclusion}

We have proven that outlier detection in motion estimation problems can benefit from the use of robust decomposition techniques, even for those cases which are beyond the theoretical limits of the robust recovery. We have exploited the information available in deficient recoveries to infer which columns are outliers. To this end, we have proposed a new method called RD-CR, able to impose rank constraints arising in motion estimation problems, in an efficient way. Additionally, we presented a robust regression framework that combines RD-CR as a preprocessing stage, prior to a fast compressed least-squares algorithm based on data reduction. We verified that this conception alleviates known problems of compressed least-squares methods and leads to a robust estimator that is both precise and fast. We have validated our approach by performing tests in both synthetic and real data sets, providing a good view of the theoretical behaviour of the method and its good performance in real scenarios. As future improvements, we are experimenting with new ways of addressing rank constraints that have proven to be computationally more efficient and capable of dealing with large-scale problems.

%% file: sections/Ack.tex
\section*{Acknowledgment}
\noindent This work has been supported by the Universitat Aut\`{o}noma de Barcelona, the Fundaci\'{o}n S\'{e}neca 08814PI08, the Spanish government, by the projects FIS201129813C0201; TRA201129454C0301 (eCo-DRIVERS) and NICTA. NICTA is an entity funded by the Australian Government as represented by the Department of Broadband, Communications, and the Digital Economy, and the ARC through the ICT Centre of Excellence Program.
\vspace{-3mm}